%% file: manuscript.tex
\documentclass[11pt]{article}
\usepackage[margin=2.4cm]{geometry}
\usepackage[T1]{fontenc}
\usepackage{lmodern}
\usepackage{microtype}
\usepackage{booktabs}
\usepackage{tabularx}
\usepackage{array}
\usepackage{amsmath}
\usepackage{amssymb}
\usepackage{url}
\usepackage[numbers,sort&compress]{natbib}
\usepackage{fancyvrb}
\usepackage{upquote}
\usepackage{graphicx}
\usepackage{capt-of} %
\usepackage{placeins}
\usepackage{tikz}
\usetikzlibrary{arrows.meta}
\usepackage[hidelinks]{hyperref}
\newcommand{\doi}[1]{\begingroup\def\UrlBreaks{\do\/\do\-}\href{https://doi.org/#1}{doi:\,\nolinkurl{#1}}\endgroup}

\newcolumntype{L}{>{\raggedright\arraybackslash}X}
\renewcommand{\arraystretch}{1.15}

\newcommand{\compacttable}{%
  \setlength{\tabcolsep}{3pt}%
  \setlength{\aboverulesep}{1.5pt}%
  \setlength{\belowrulesep}{1.5pt}%
  \setlength{\abovecaptionskip}{4pt}%
  \renewcommand{\arraystretch}{1.02}%
}

\title{
    Guessing Without Momentum:\\
    A Language Model States the Rules but Skips the Test
}
\author{
    Andrei Aioanei$^{1}$,\quad
    Jens Lehmann$^{2}$,\quad
    Sahar Vahdati$^{1,2,3}$\\[0.8em]
    {\small $^1$TIB -- Leibniz Information Centre for Science and Technology, Germany}\\
    {\small $^2$Dresden University of Technology, Germany}\\
    {\small $^3$L3S Research Center, Leibniz University of Hannover, Germany}\\[0.3em]
    {\small\texttt{Andrei.Aioanei@tib.eu}\quad
    \texttt{jens.lehmann@tu-dresden.de}\quad
    \texttt{sahar.vahdati@tib.eu}}\\[0.3em]
    {\small Team NIMI in the ARC Prize 2026, \url{https://nimi-ai.com}}
}
\date{}

\begin{document}
\maketitle

\input{sections/00-abstract}
\input{sections/01-introduction}
\input{sections/02-setting}
\input{sections/03-model-alone}
\input{sections/04-harness-help}
\input{sections/05-why-plays-stop}

\input{sections/08-related-work}
\input{sections/11-conclusion}

\section*{Ethics statement}
The eighteen LS20 trajectories analysed here were recorded during voluntary
workshop and seminar sessions and were first reported in \citet{ABA+26}.
Participants were informed that their gameplay data could be used for
research purposes. The recordings hold actions, boards and timestamps under
aliases generated by the game portal and contain no participant names.

\section*{Data and code availability}
The ARC-AGI-3 games are available at \url{https://three.arcprize.org}. The
notebook of our entry is public at
\url{https://www.kaggle.com/code/andreicaioanei/semiotic-scaffold-r41}.

\section*{Acknowledgements}
We thank the workshop participants, and the ARC Prize Foundation for making
ARC-AGI-3 available for research.

\clearpage
\bibliographystyle{arccompact}
\bibliography{references}

\appendix
\input{sections/A-what-the-model-is-sent}
\input{sections/B-stopped-plays}
\input{sections/C-related-work-in-full}
\input{sections/D-further-measurements}

\end{document}

%% file: sections/00-abstract.tex
\begin{abstract}
A language-model agent can reach a goal in an unfamiliar environment while its account
of the rules remains inadequate for later levels. We study this gap with a quantised
Qwen3.8-Flash-Next model in ARC-AGI-3, an interactive benchmark whose rules and goals
must be discovered through play. The philosopher Charles Sanders Peirce described
inquiry as a cycle of proposing explanations, deriving predictions, and testing them
against experience. We use these stages to analyse recorded plays, changes to what the
model observes and which tools it has, and preliminary fine-tuning.
To test a proposed fix, we restore a game and the model's conversation at a chosen
decision point and let the model continue repeatedly, with and without the fix, under
the same token budget. At four points from failed plays, 19 of 40 continuations without
any fix still completed the level, so a fix must be compared with this unaided recovery.
Computed summaries of the game added to every turn coincided with fewer completed
levels. Factual notes improved the next response, but a broader correction package
produced no detectable gain in score. A cue to an intermediate board the model had not
been shown coincided with higher completion at selected points in one game, with no detected benefit
at three others. Fine-tuning made the model check its rules far more often and write
longer replies, but it completed fewer held-out levels under the same time limits. More checking did not mean better play, which raises the question of when a check is worth its cost.
Human and model traces motivate \emph{momentum in inquiry}: continuing to investigate an
unresolved relation even when the goal can already be reached. We propose a drive to
understand as an explanation of the human pattern. Whether such inquiry improves rule
revision and transfer remains to be tested.
\end{abstract}

%% file: sections/01-introduction.tex
\section{Introduction}

ARC-AGI-3 asks an agent to play unfamiliar games without being told their rules or
goals \citep{ARC26,arcPrize2026Competition}. To progress reliably, the agent must learn
what its actions control, which changes matter, and what completes a level. In a
language-model agent, the model brings general abilities to interpret images, write
programs and plan, while the surrounding \emph{harness} controls its observations, tools
and memory.

Completing a level does not show that the rule behind it is adequate for what follows:
success can leave competing explanations unresolved, and performance alone does not
establish the understanding attributed to it \citep{mitchell2026sixPrinciples}.
Conversely, a change to the harness can correct the model's next response, or make it
check more, without improving completion: added descriptions use up context, checks use
up computation, and exploratory actions lower the action-efficiency score. We therefore
ask: \emph{when do changes in how evidence is presented, or in how hypotheses are
tested, lead a resource-bounded language model to complete more of ARC-AGI-3?}

\begin{figure}[!ht]
    \centering
    \resizebox{\textwidth}{!}{\input{figures/inquiry-ls20}}
    \caption{Peirce's terms illustrated on LS20. Board elements are redrawn from a screenshot, and the transition and rival hypotheses are schematic. Sign types (icon,
    index, symbol) differ from inquiry stages.}
    \label{fig:inquiry-ls20}
\end{figure}
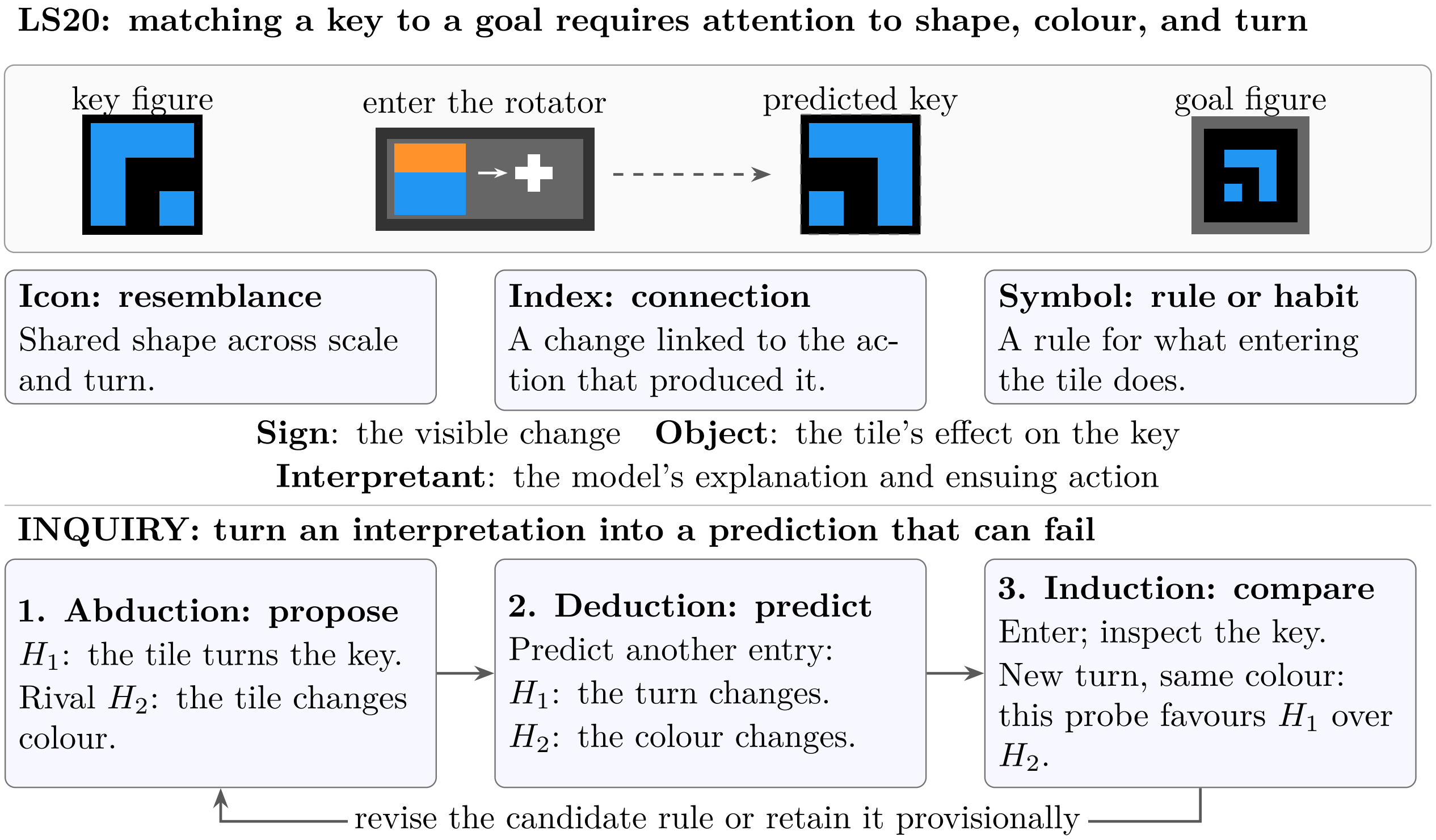

We study a quantised Qwen3.8-Flash-Next model in a Python-based harness derived from a
published competition entry \citep{bessisEtAl2026duck,franzen2026milestone2}, using 100
published plays of 25 public games, further experiments, an analysis of failed plays,
and fine-tuning. Our vocabulary is Charles Sanders Peirce's account of inquiry
\citep{Pei31,Pei98,Sho07}: \emph{abduction} proposes a candidate rule, \emph{deduction}
derives what it predicts, and \emph{induction} tests the predictions against
experience. We call the disposition to carry out this cycle a \emph{habit of inquiry}. Figure~\ref{fig:inquiry-ls20} illustrates these terms on the game LS20.

Our comparison of human and model play (Section~\ref{sec:momentum}) asks whether a
player keeps investigating an event when a route to the goal is already available.
We call this sustained investigation \emph{momentum in inquiry}. In the example,
the human returns to the tile and gains another observation of its effect, while
the model reaches the goal without resolving competing explanations. A drive to
understand is one possible explanation for the human's return.

To evaluate a fix, we restore a failed play at a chosen decision point, with its board
and conversation, and let the model continue ten times with and ten times without the
fix under the same token budget (Section~\ref{sec:instrument}). Without any fix, 19 of
40 continuations still completed the level, so comparing a fix with the recorded failure
alone would overstate its effect. We then test whether harness changes and fine-tuning that
alter how the model uses evidence or checks its rules improve later play. We contribute
this controlled evaluation of fixes and a distinction, grounded in recorded play, between
reaching a goal and testing the rule used to reach it. The lessons that carry over
concern unaided recovery, checks that tell rival rules apart, and completion within
budget.
This work accompanies Team NIMI's entry on the Kaggle ARC Prize leaderboard: the
\href{https://www.kaggle.com/code/andreicaioanei/semiotic-scaffold-r41}{r41 notebook}
(submission \texttt{57727775}) scored 34.38, the best of nine runs, which averaged 29.4
and varied by about three points.

%% file: figures/inquiry-ls20.tex
\begin{tikzpicture}[
  panel/.style={draw=black!55, rounded corners=3pt, fill=blue!3,
    inner sep=4pt, font=\small, align=left},
  arr/.style={-{Stealth[length=6pt]}, thick, black!65},
  lsblue/.style={fill={rgb,255:red,33;green,150;blue,243}, draw=none}]
\node[anchor=west, font=\small\bfseries] at (0,0.2)
  {LS20: matching a key to a goal requires attention to shape, colour, and turn};
\draw[rounded corners=3pt, black!40, fill=black!2] (0,-0.25) rectangle (15,-2.22);
\node[font=\small] at (1.45,-0.63) {key figure};
\fill[black] (0.82,-2.03) rectangle (2.08,-0.77);
\foreach \x/\y in {0/0,1/0,2/0,0/1,0/2,2/2}
  {\fill[lsblue] ({0.91+0.36*\x},{-0.86-0.36*\y})
    rectangle ++(0.36,-0.36);}
\node[font=\small] at (5.05,-0.63) {enter the rotator};
\fill[black!80] (3.9,-1.99) rectangle (6.2,-0.91);
\fill[black!60] (4.02,-1.87) rectangle (6.08,-1.03);
\fill[lsblue] (4.10,-1.075) rectangle (4.85,-1.825);
\fill[fill={rgb,255:red,255;green,146;blue,43}]
  (4.10,-1.075) rectangle (4.85,-1.375);
\foreach \x/\y in {0/0,-1/0,1/0,0/-1,0/1}
  {\fill[white] ({5.50+0.13*\x},{-1.45+0.13*\y}) rectangle ++(0.13,0.13);}
\draw[-{Stealth[length=4pt]}, white, thick] (4.98,-1.385) -- (5.29,-1.385);
\node[font=\small] at (9,-0.63) {predicted key};
\draw[black!55, dashed, fill=black] (8.37,-2.03) rectangle (9.63,-0.77);
\foreach \x/\y in {0/0,1/0,2/0,2/1,0/2,2/2}
  {\fill[lsblue] ({8.46+0.36*\x},{-0.86-0.36*\y})
    rectangle ++(0.36,-0.36);}
\draw[arr, dashed] (6.4,-1.4) -- (8.06,-1.4);
\node[font=\small] at (13.1,-0.63) {goal figure};
\fill[black!60] (12.48,-2.03) rectangle (13.72,-0.79);
\fill[black] (12.61,-1.9) rectangle (13.59,-0.92);
\foreach \x/\y in {0/0,1/0,2/0,2/1,0/2,2/2}
  {\fill[lsblue] ({12.83+0.18*\x},{-1.14-0.18*\y})
    rectangle ++(0.18,-0.18);}
\node[panel, text width=4.25cm, minimum height=1.3cm, anchor=north west] at (0,-2.4)
  {\textbf{Icon: resemblance}\\[1pt]
   Shared shape across scale and turn.};
\node[panel, text width=4.25cm, minimum height=1.3cm, anchor=north west] at (5.15,-2.4)
  {\textbf{Index: connection}\\[1pt]
   A change linked to the action that produced it.};
\node[panel, text width=4.25cm, minimum height=1.3cm, anchor=north west] at (10.3,-2.4)
  {\textbf{Symbol: rule or habit}\\[1pt]
   A rule for what entering the tile does.};
\node[font=\small, align=center, text width=14.6cm] at (7.5,-4.37)
  {\textbf{Sign}: the visible change\quad
   \textbf{Object}: the tile's effect on the key\\[1pt]
   \textbf{Interpretant}: the model's explanation and ensuing action};
\draw[black!25] (0,-4.88) -- (15,-4.88);
\node[anchor=west, font=\small\bfseries] at (0,-5.16)
  {INQUIRY: turn an interpretation into a prediction that can fail};
\node[panel, text width=4.25cm, minimum height=2.4cm, anchor=north west] (a) at (0,-5.44)
  {\textbf{1. Abduction: propose}\\[1pt]
   $H_1$: the tile turns the key.\\[1pt]
   Rival $H_2$: the tile changes colour.};
\node[panel, text width=4.25cm, minimum height=2.4cm, anchor=north west] (d) at (5.15,-5.44)
  {\textbf{2. Deduction: predict}\\[1pt]
   Predict another entry:\\[1pt]
   $H_1$: the turn changes.\\[1pt]
   $H_2$: the colour changes.};
\node[panel, text width=4.25cm, minimum height=2.4cm, anchor=north west] (i) at (10.3,-5.44)
  {\textbf{3. Induction: compare}\\[1pt]
   Enter; inspect the key.\\[1pt]
   New turn, same colour: this probe favours $H_1$ over $H_2$.};
\draw[arr] (a.east) -- (d.west);
\draw[arr] (d.east) -- (i.west);
\draw[arr] (i.south) -- ++(0,-0.35) -| (a.south);
\node[fill=white, font=\small, inner sep=2pt] at (7.5,-8.19)
  {revise the candidate rule or retain it provisionally};
\end{tikzpicture}

%% file: sections/02-setting.tex
\FloatBarrier
\section{Model and evaluation setting}
\label{sec:setting}
\label{sec:harness}

We study a quantised Qwen3.8-Flash-Next model in Franzen's published version of
Tufa Labs' Duck harness \citep{bessisEtAl2026duck,franzen2026milestone2}. Our entry
uses the same weights, adds switchable observation changes and runs a modified inference
server (Appendix~\ref{app:loop}). Appendix~\ref{app:configurations} maps each study's
configuration to the submitted notebook.

Each turn, the harness sends the model an \emph{opener}: the actions just played and
images of the current board and of the changed cells. The model replies with reasoning
and Python code that inspects the game record or plays one or several actions. Functions it defines stay available. The record keeps every action with its resulting board,
including intermediate boards the images do not show. The model can also read a
\emph{percept}, the harness's description of the board as single-colour patches with
their shapes and the patches they enclose or touch. Conversation history is cut to half
at about 118,000 estimated tokens, while the game record is kept in full.

Our baseline data are the published plays of this harness, four on each of 25 public games. Other datasets and budgets are given with their experiments. The competition runs
110 hidden games on one RTX PRO 6000 GPU with 96 GB of memory, within nine hours and
without network access \citep{kaggle2026arcagi3}. Ten games share the model server at a
time, giving each game about 48 minutes of service on average. We call runs under these conditions \emph{full-length}. Completing levels is not the same as the competition
score, which also rewards action efficiency and weights later levels more heavily
\citep{arcprize2026methodology}. We report public and hidden games separately.

%% file: sections/03-model-alone.tex
\section{What the model does by itself}
\label{sec:alone}

We first examine the 100 published plays of the original harness, plus further runs
named below. The percept (Section~\ref{sec:setting}), the harness's list of the board's
single-colour patches and of which enclose or touch others, describes only the board.
The model must still work out how actions change it and what completes a level, and
its failures suggest targets for the help in
Section~\ref{sec:help}.

\subsection{One game: human and model trajectories}

In LS20, the player moves to a goal that admits them only when the key, a figure in
the lower left corner, matches the goal's figure in shape, colour and orientation (six,
four and four options). Floor tiles turn the key or change its colour or
shape, and each level adds something new (Table~\ref{tab:ls20levels}).

To see exactly what each player did, we re-ran on the game's engine, that is, the game's
own code, the recorded actions of four published model plays and of the seven human
recordings made with the current game version (Appendix~\ref{sec:people}). The engine's
state shows which tile each action touched and how the key changed, without interpreting screenshots. The re-run reproduced the recorded boards in full for
all but two human recordings. Six human recordings complete all seven levels, while the four model plays end on levels 4, 5, 5 and 6. Table~\ref{tab:ls20people} compares both groups'
actions and tile contacts.

\begin{table}[ht]
    \centering
\compacttable
\caption{LS20 on the game's engine: actions per level and contacts with key-changing tiles,
for six human recordings and the model plays completing the level. ``Steps needed'':
required key changes.}
    \label{tab:ls20people}
    \small
    \begin{tabularx}{\textwidth}{cLLlLL}
    \toprule
Level & Actions, human & Actions, model & Steps needed & Steps, human
      & Steps, model \\
    \midrule
    1 & 13, 15, 15, 17, 23, 39 & 21, 33, 40, 46 & turn 1 & 1, 1, 1, 1, 1, 4
      & 1, 5, 5, 9 \\
    2 & 45, 45, 47, 67, 71, 475 & 45, 45, 45, 107 & turn 3 & 3, 3, 3, 6, 6, 26
      & 3, 3, 3, 7 \\
    3 & 43, 50, 55, 79, 79, 128 & 78, 88, 90, 160 & turn 2, colour 1
      & turn 2, 2, 2, 2, 4, 5 and colour 1, 1, 2, 2, 3, 5
      & turn 4, 4, 4, 6 and colour 1, 1, 1, 3 \\
    4 & 50, 54, 54, 56, 76, 85 & 66, 103, 111 & colour 3, shape 1
      & colour 3 and shape 1 in all six & colour 3, 4, 14 and shape 1, 2, 7 \\
    5 & 49, 53, 58, 59, 98, 144 & 152 & shape 2, colour 3, turn 2
      & colour 3 in all six, shape 2 in five & shape 9, colour 4, turn 19 \\
    \bottomrule
    \end{tabularx}
\end{table}

Noticing a mechanic, understanding it and using it are separate steps. At the start of
level 3, all four model plays name the new colour tile, and one calls it a ``color wheel''.
All seven human recordings step on the colour tile before the turning tile, whereas
three model plays first walk to the goal with the wrong colour and lose a life. Each
also hit, mid-batch, a strip that throws the player to the next wall. The model blamed a broken position function, a teleport or moving walls, and spent five to nine replies
(9,000--18,000 tokens) on it while keeping its plan.

Precision matters too. On level 1, where one turn of the key suffices, two model plays
turn it five and nine times, having mistaken the key's match with the goal for
completion. A level-3 play plans its route with a rule for the move bar that contradicts
some of its own printed measurements (Section~\ref{sec:critique}). The model can thus
identify the relevant objects and state plausible rules without an account accurate
enough to guide its play.

\subsection{Momentum in inquiry: an event changes what matters}
\label{sec:momentum}

By \emph{momentum in inquiry} we mean that an unresolved relation keeps priority in a
player's next steps: returning to it, probing it, revising and acting. A pause after an event matters when it changes what the player
investigates next, reorganizing which explanations and actions seem worth considering.

\input{sections/03-momentum-comparison}

Human timing motivates this account: across eighteen completed LS20 trajectories, the
median pause after an action that changed the key was 0.96 seconds, against 0.46 after
other actions on the same level. For the model we can compare only reply lengths, which
do not measure deliberation in the same way (Appendix~\ref{sec:people}). We hypothesize
that such pauses serve \emph{online hypothesis pruning}: discarding contradicted accounts
and choosing probes that separate the rest.

\subsection{What its programs read and what they work out}

Figure~\ref{fig:program-use} counts code patterns in the 9,170 distinct programs of the
100 published plays. The model rotates cells in its own code more often than it reads
the rotation-invariant shape signature that the percept supplies. Low use alone does not
show whether a tool is unsuitable or merely unnecessary. LS20 shows one concrete
mismatch: matching key and goal requires comparing size, colour, orientation and
diagonally touching parts, which a signature of single-colour patches does not provide.

\begin{figure}[!ht]
\centering
\resizebox{\textwidth}{!}{\input{figures/program-use-compact}}
\caption{How the model's programs use the board (\% of 9,170 programs): \textcolor[RGB]{39,108,155}{\rule{.9em}{.55em}}\,reading data, \textcolor[RGB]{191,91,37}{\rule{.9em}{.55em}}\,the model's own computations, \textcolor[RGB]{36,123,117}{\rule{.9em}{.55em}}\,reading percept relations. 44.5\% read neither the letter board nor the percept patches, working only from retained data or actions. No percept relation is read in over 2.1\%.}
\label{fig:program-use}
\end{figure}

To examine the model's own code, we read forty \emph{stays}, the part of a play spent on
one level, where another play completed that level. Thirteen contained an incorrect
measurement, search or simulator, and in one the problem turned on a percept relation. These
diagnoses suggest fixes without showing that they would help. Appendix~\ref{app:carry} relates code reuse across levels to completion.

%% file: sections/03-momentum-comparison.tex
\begin{figure}[!ht]
\centering
\includegraphics[trim=0 480bp 0 0,clip,width=0.76\textwidth]{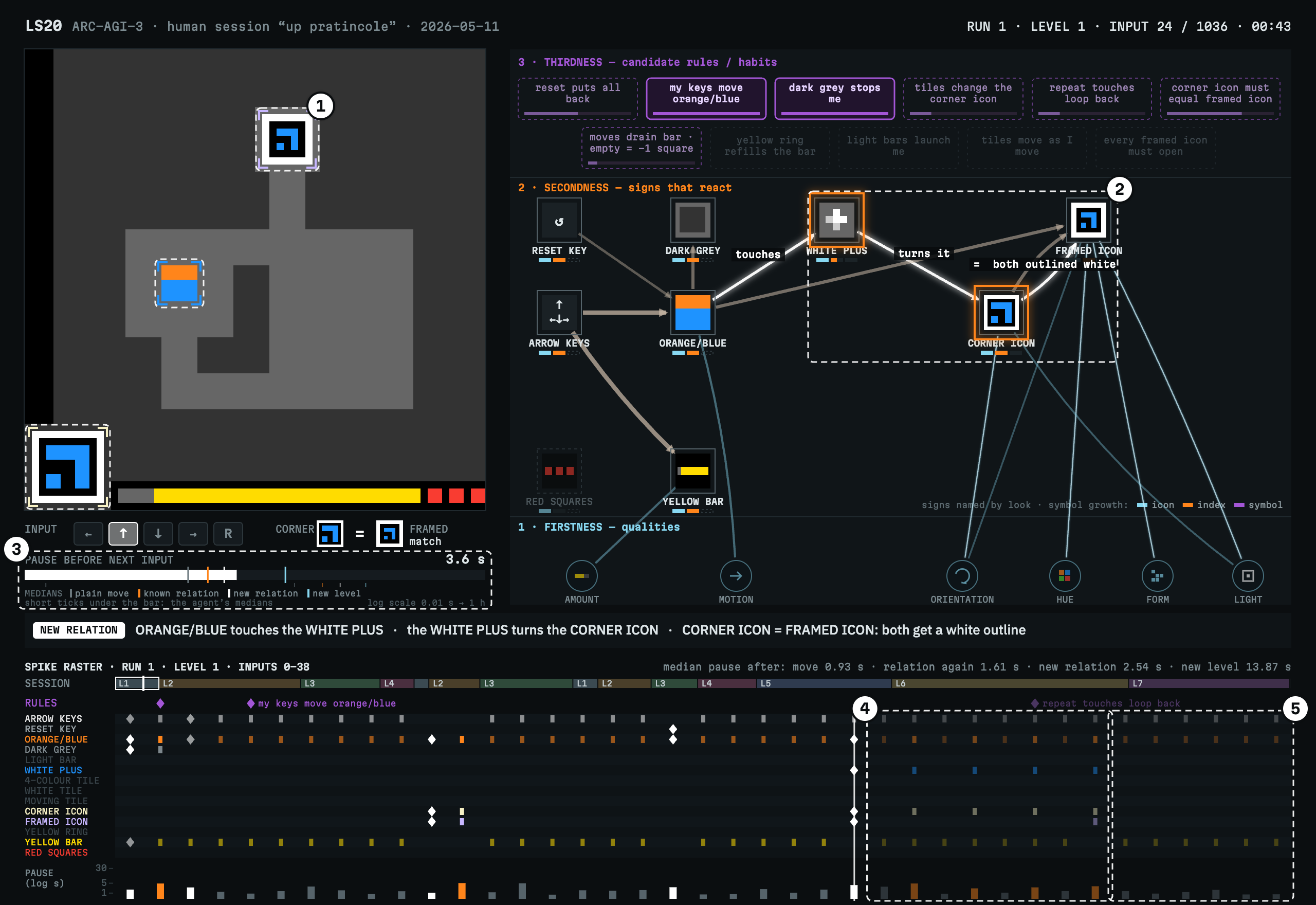}\\[2pt]
\includegraphics[trim=0 480bp 0 0,clip,width=0.76\textwidth]{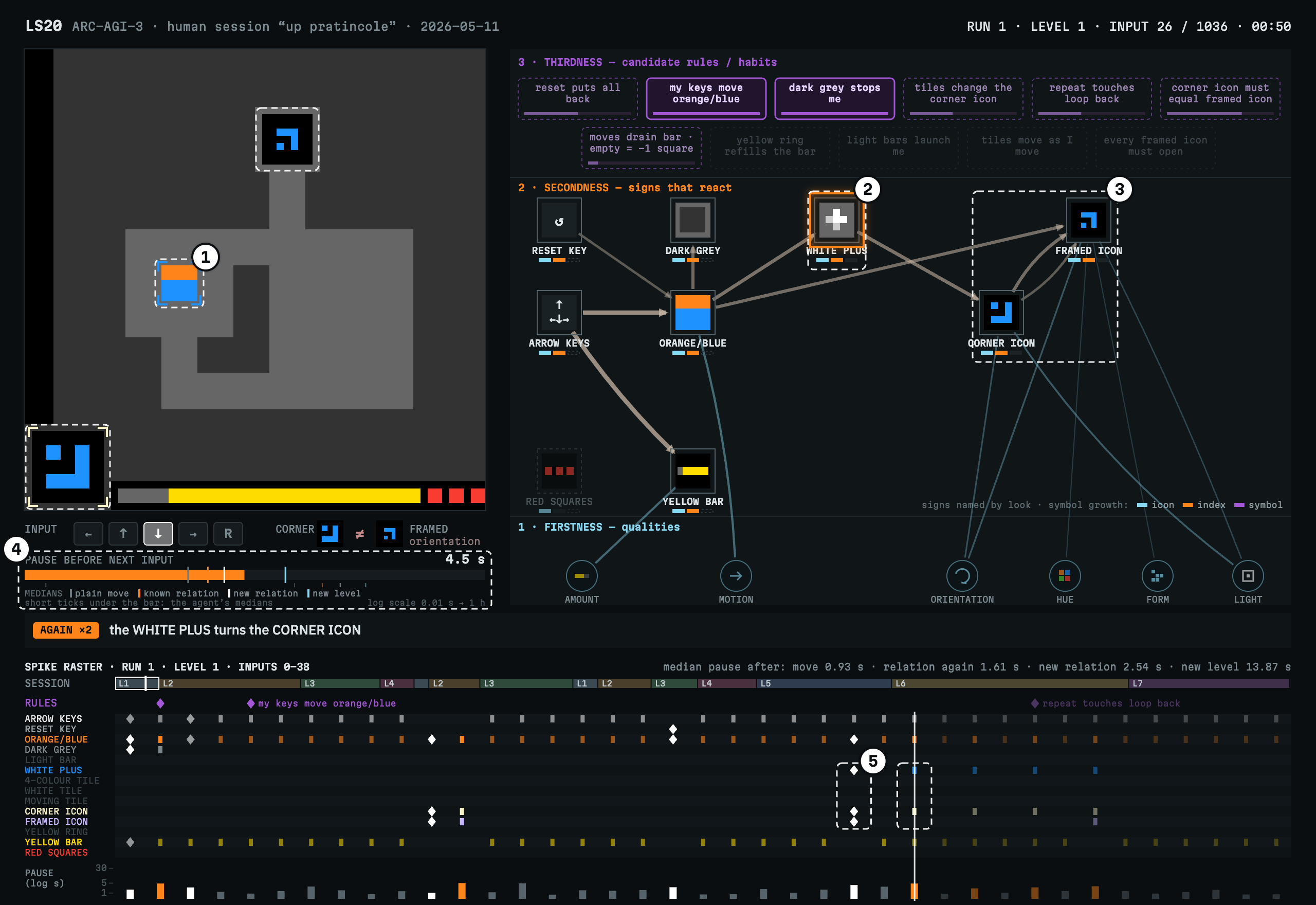}
\caption{A human playing LS20. Top: at input~24 the player steps on the rotator, the key
now matches the goal, and the player pauses 3.6\,s. Bottom: at input~26 the player steps
on it again, undoing the match, and pauses 4.5\,s. The relation maps illustrate how
linked events can form a concept. They do not measure the player's concepts or purpose.}
\label{fig:momentum-comparison}
\label{fig:momentum-h1}
\label{fig:momentum-h1b}
\end{figure}

\begin{figure}[!ht]
\centering
\includegraphics[trim=0 528bp 0 0,clip,width=0.76\textwidth]{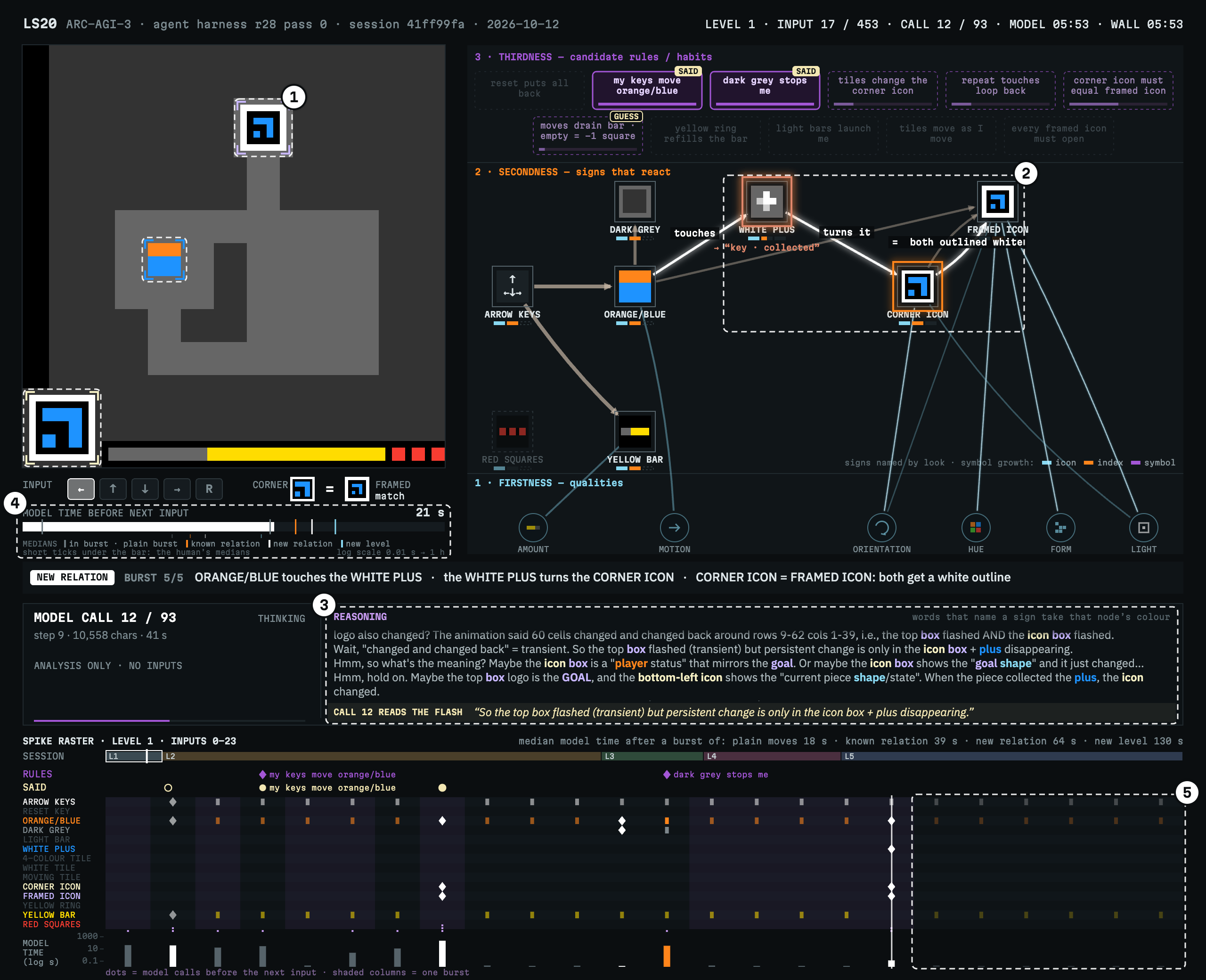}\\[2pt]
\includegraphics[trim=0 528bp 0 0,clip,width=0.76\textwidth]{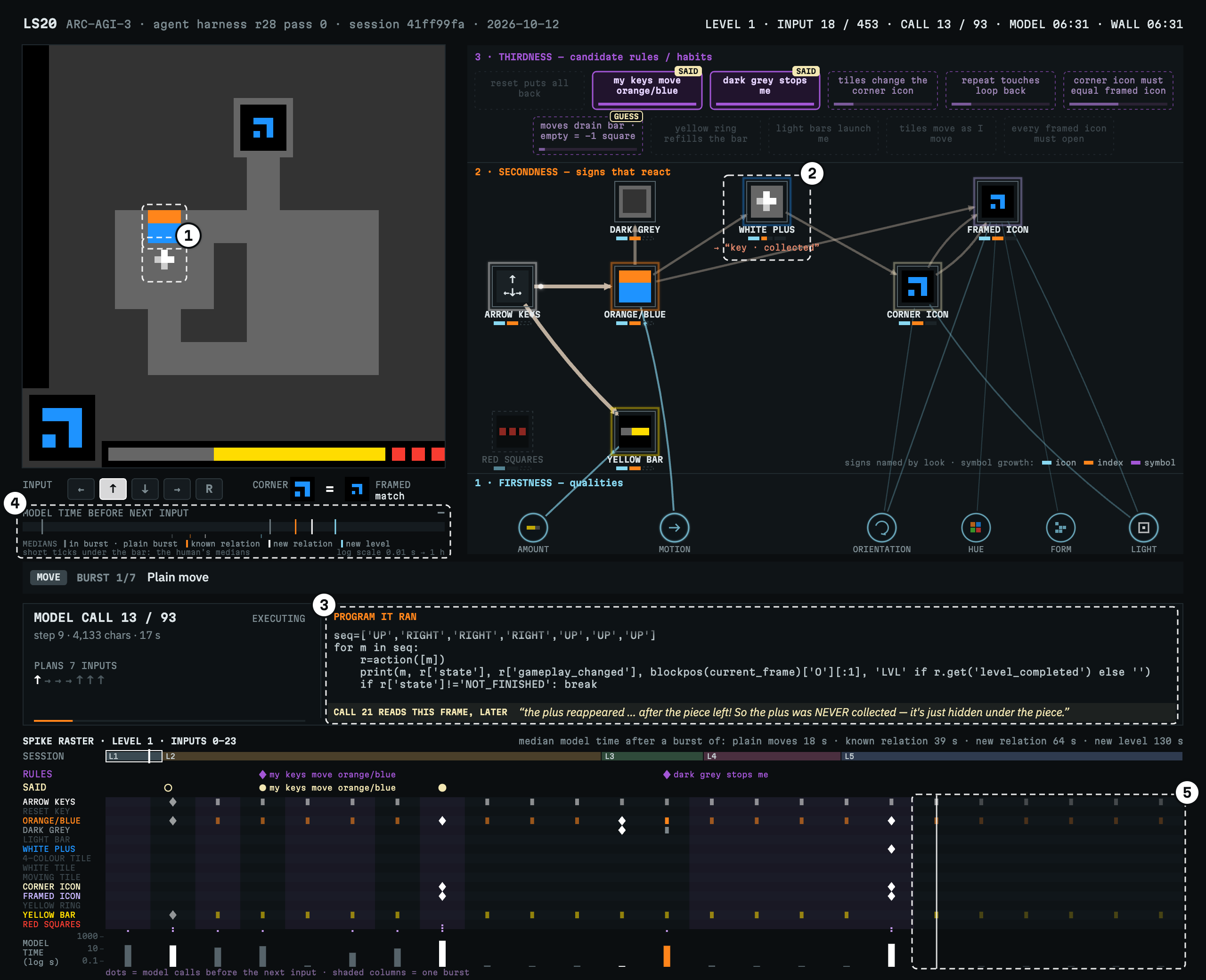}
\caption{Our model playing LS20, in the first full-length play of our earlier entry r28.
Top: at input~17 (model call~12) the model notices the
flash that signals the key--goal match. Bottom: call~13 starts a seven-input route to the
goal at input~18, and the excerpt includes the model's correction, at call~21, of its belief
that the plus had been collected. The maps separate events from the model's stated accounts. Progress does not validate its explanation.}
\label{fig:momentum-a1}
\label{fig:momentum-a1b}
\end{figure}

Figures~\ref{fig:momentum-comparison} and~\ref{fig:momentum-a1} follow one human
session and one model play around the moment the key first matches the goal. Their
labels use Peirce's three categories: \emph{Firstness} for qualities such as a colour,
\emph{Secondness} for actual encounters and reactions, and \emph{Thirdness} for the rules
and habits that mediate interpretation \citep{Pei31,Sho07}.
At input~26 (Figure~\ref{fig:momentum-comparison}, bottom), the human steps back onto
the rotator, a white plus, which turns the key again and undoes the match. This return
yields another observation of how the rotator changes the key, which walking into the
goal would not have provided. It is common: after the key's first turn, 17 of 18 human
trajectories paused longer than usual, and 9 turned the key again before entering the
goal. We read such linked events as \emph{concept formation}: separate changes become a
candidate causal relation that keeps organizing play.

At call~12 (Figure~\ref{fig:momentum-a1}, top), the model notices the flash but takes
the plus to have been collected. Call~13 begins a seven-input route to the goal
(Figure~\ref{fig:momentum-a1}, bottom). In this state, both the collection account and
the actual key--goal match predict that the goal will open, so reaching it cannot tell
them apart. At call~21 the model corrects itself: ``the plus was NEVER collected''. So the model does revise. The question is whether the unresolved relation then directs what it
probes and does. Of the four full-length LS20 plays of our earlier entry r28, one turned
the key again.

The sequences thus contrast another observation of the relation with a successful
action consistent with both accounts. Playfulness or a drive to understand may sustain
the human's inquiry, but the sequences establish no motive: a return can serve other
purposes, and a goal-directed action can be a probe.

%% file: figures/program-use-compact.tex
\begin{tikzpicture}[x=1cm,y=1cm,font=\footnotesize]
\definecolor{programaccess}{RGB}{39,108,155}
\definecolor{programcompute}{RGB}{191,91,37}
\definecolor{programpercept}{RGB}{36,123,117}
\foreach \v in {0,10,20,30,40,50} {
 \draw[black!12] ({8+.14*\v},.15) -- ({8+.14*\v},-3.82);
 \node[anchor=north] at ({8+.14*\v},-3.84) {\v};
}
\foreach \label/\share/\yy/\tone in {
 {Reads board as letters}/38.4/0/programaccess,
 {Reads percept patches}/22.3/-.34/programaccess,
 {Reads neither; uses retained data / actions}/44.5/-.68/programaccess,
 {Counts a bar or colour}/16.7/-1.18/programcompute,
 {Builds a cell lattice}/8.6/-1.52/programcompute,
 {Searches a path}/7.1/-1.86/programcompute,
 {Turns / flips cells}/2.2/-2.20/programcompute,
 {Reads enclosure}/2.1/-2.70/programpercept,
 {Reads contact}/0.7/-3.04/programpercept,
 {Reads shape signature}/0.4/-3.38/programpercept} {
 \node[anchor=east] at (7.8,\yy) {\label};
 \fill[\tone] (8,{\yy-.105}) rectangle ({8+.14*\share},{\yy+.105});
 \node[anchor=west,inner sep=3pt] at ({8+.14*\share},\yy) {\share\%};
}
\end{tikzpicture}

%% file: sections/04-harness-help.tex
\section{What a harness can give}
\label{sec:help}

If the model fails because it misses evidence, the harness should be able to help by
supplying it. We tried three kinds of help, each aimed at one way the model can miss
evidence. If it never computes a relation, we \emph{tell} it computed summaries. If it
misreads a relation it did compute, we \emph{correct} the fact. If it never sees a board,
we \emph{show} it with a cue. The model often took up a specific fact it was given, and
its play improved at VC33, where the cue named a colour the recorded play had missed.
Broader help brought no gain or cost levels.

\subsection{Relational summaries}
\label{sec:telling}

Summaries describing the game did not help, and they used up the space that the
model's own history needed. A module added a computed summary to every turn: effects of
recent actions, event frequencies, recurring changes, blocked moves, similar figures and
how the previous level was completed. It described the game but set no goals and
recommended no actions. On the competition GPU, ten-game runs completed 43, 42 and 41
levels with summaries and 45 and 45 without (hosted runs in
Appendix~\ref{app:harness-comparisons}). On the same games and server, the counts were
41 versus 45, and the history had to be cut 20 versus 13 times. Late in a game the
summaries reportedly made up 58\% of the turn message. Displaced context, overlapping
descriptions and serving costs are possible explanations. Whatever the mechanism, in a
fixed context every line the harness adds displaces something the model wrote itself,
so help has to earn its space.

\subsection{Targeted factual corrections}
\label{sec:correcting}

In most of the costly errors we examined, the evidence was already there. We chose eight
levels on which one play was far more costly than another. In six, the costlier play had found a useful rule and then misread or miscompared
evidence that was already in its own tool output.

\begin{figure}[ht]
\centering
\includegraphics[width=\textwidth]{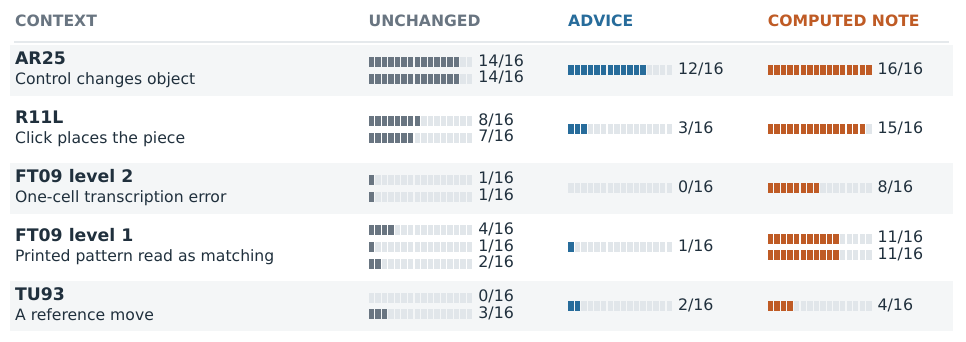}
\caption{Agreement of the next step with the successful reference play. Each strip is one
batch of 16 replies to the same request, and filled cells agreed. Batches are unpaired across conditions.}
\label{fig:correction-batches}
\end{figure}

\noindent\textbf{Next-response probes.} At the point of the error, we sampled replies to
the recorded request unchanged, with five sentences of advice to check, or with a
computed note stating the missed fact. A reply matched if its next step agreed with what
a successful reference play did there, a criterion that ignores other valid actions and
later completion. Computed notes increased matches, whereas checking advice did not
(Figure~\ref{fig:correction-batches}). The model can use a fact when it is handed over, but being told to check does not make
it find the fact itself.

\noindent\textbf{Whole-game package.} In whole games, a broader package brought no
detectable gain. It combined such notes with action probes, tracebacks of failing
functions and a reset the model could request. Each arm played the 25 public games 39
times (975 plays, 73 million generated tokens). Mean scores were 18.84 without and 19.19 with the package (difference 0.35, reported
standard error 0.67),
and the lead came mostly from two games used in development. This gap between a better
next step and a better game motivates Section~\ref{sec:instrument}.
Appendices~\ref{app:intervention-details} and~\ref{app:harness-comparisons} cover wording and hidden-game scores.

\subsection{Controlled continuations}
\label{sec:instrument}

A diagnosed error suggests a fix, but a single replay with the fix cannot show that it
helped: the model's replies are sampled, and another attempt might have succeeded anyway.
We therefore replay a recorded play on the game's engine up to a chosen request,
restoring the board and the model's conversation, and sample ten continuations with the
fix and ten without. Each ends when the level is completed or a token budget, equal
across arms, runs out, which counts as failure. Continuations without the fix measure unaided
recovery. We record completion, tokens and actions, with outcomes, thresholds and
decision rules fixed in writing before each round.
The estimates hold for the chosen point and budget, and equal tokens need not mean equal
wall time. Complementing evaluations of corrections and of whole harnesses
\citep{tyen2024mistakes,rodionov2026ablation,lehmannVahdati2026harnessImprovement}, the
method judges a diagnosed fix by its effect on completion, and it needs only recorded
model requests and a replayable environment.

\subsection{Intermediate-board cues}
\label{sec:hidden}

\begin{figure}[ht]
\centering
\includegraphics[width=\textwidth]{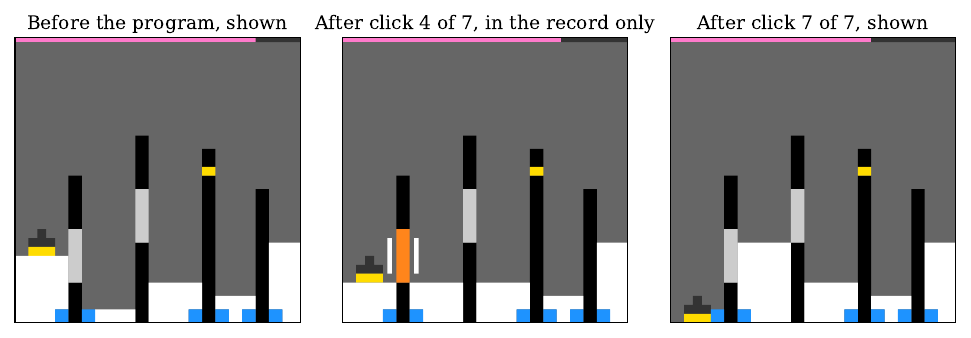}
\caption{VC33 level 4, replayed on the game's engine. The fourth click made the bar
orange and the fifth made it light grey. Only the record kept the orange state, which neither default image showed.}
\label{fig:vc33}
\end{figure}

\noindent\textbf{Cue and design.} In most turns, some boards the model produced are
never shown to it. In 2,832 of the 4,434 turns of the published plays, the model plays
several actions, and the boards between them are kept in \texttt{history} but need not
appear in the next opener's two images. On VC33 level~4, the fourth of seven
clicks turned a 36-cell bar orange and the fifth turned it grey again
(Figure~\ref{fig:vc33}). The model omitted this bar from its measurements, mentioned orange 32 actions and
67,000 tokens later, and lost the level.

A colour held only between two openers triggers a cue in the next one, at most once
per colour and level:
\begin{quote}
\small
Between your previous analyzer turn and now the board held a color that neither the
previous board nor the current one shows: after action 4 of 7 (\texttt{MOUSE(row=62,
col=16)}) 36 cells were O (orange) around rows 43-54, cols 12-14.
\texttt{history[-4].frame} is that board.
\end{quote}
The cue states a fact and, as a Peircean \emph{index}, points to the board that shows it
\citep{Sho07}. It is rare, firing in 48 of 4,510 openers of recorded plays (28 plays, 11
games), and extends the published harness's report of colours seen only during one
action's animation \citep{franzen2026milestone2} to boards between a turn's actions.

Round~1 switched on both the colour cue and a cue about an earlier animation, at two
decision points in VC33, two in LS20 and one in LF52. Round~2 used the colour cue alone,
at the first VC33 point and at points from failed plays in DC22, WA30 and M0R0.
\begin{table}[ht]
\centering
\compacttable
\caption{Completions out of ten continuations per arm and decision point, without (plain)
and with the cue. Medians of tokens and actions cover successful continuations only.}
\label{tab:rounds}
\small
\begin{tabularx}{\textwidth}{Lrrrr}
\toprule
Context & Complete, plain & Complete, cue & Tokens/actions, plain & Tokens/actions, cue \\
\midrule
\multicolumn{5}{l}{\emph{Round 1: colour-cue contexts}} \\
VC33 level 4, first play & 4/10 & 9/10 & 27,300 / 44.5 & 13,264 / 24 \\
VC33 level 4, third play & 4/10 & 7/10 & 24,931 / 25 & 24,861 / 26 \\
\multicolumn{5}{l}{\emph{Round 1: earlier-animation contexts}} \\
LS20 level 3, first play & 6/10 & 7/10 & 24,930 / 76.5 & 25,000 / 46 \\
LS20 level 3, second play & 6/10 & 9/10 & 26,453 / 63.5 & 25,001 / 71 \\
LF52 level 2 & 1/10 & 0/10 & -- & -- \\
\multicolumn{5}{l}{\emph{Round 2: colour cue alone}} \\
VC33 level 4, first play & 5/10 & 8/10 & 15,161 / 33 & 15,650 / 28.5 \\
DC22 level 4 & 5/10 & 4/10 & 47,265 / 78 & 39,985 / 76 \\
WA30 level 2 & 4/10 & 5/10 & 42,874 / 45 & 43,586 / 42 \\
M0R0 level 3 & 5/10 & 5/10 & 35,735 / 56 & 38,788 / 55 \\
\bottomrule
\end{tabularx}

\end{table}

\noindent\textbf{What the cue changed.} The cue helped at VC33, where the recorded play
had left the orange bar out of its measurements and lost the level. At the first VC33
point, completion rose from 9/20 to 17/20 over the two rounds, and at the other VC33
point from 4/10 to 7/10 (Table~\ref{tab:rounds}). All 30 cued VC33 continuations
mentioned orange in their first reply. VC33 is also where we found the gap, so this
evidence comes from the case the cue was built for. At the three other points of
round~2, the cue was mentioned in 29/30 continuations against 9/30, yet each arm
completed 14/30 (differences of $-1$, $+1$ and zero). There, naming the colour did not
change completion. These counts do not show whether the colour mattered in those plays,
and samples this small cannot rule out a small effect.
In LS20, the cue about an earlier animation made every first reply mention the throw,
which uncued replies did by their third. Completion was 16/20 against 12/20, yet the
behaviour the cue targeted did not change: in each arm, the model visited the colour
tile before trying the goal in 8/20 continuations (Appendix~\ref{app:intervention-details}).
The model used the cue's words more than its pointer: in round~1, 39/50 cued
continuations mentioned the fact, but only 19/50 retrieved the board before
acting (2/50 uncued, target 25/50).

\noindent\textbf{Unaided recovery.} The four points of round~2 came from plays that
failed after spending 82,000--127,000 tokens on the level. Yet without any cue, 19/40
continuations completed the level within 50,000 or 60,000 tokens of the restored
request. A fix tried once from such a point would therefore look successful about half
the time even if it did nothing. A failed trace shows where one sample went wrong. It
does not by itself show what the model cannot do.

\noindent\textbf{Across the three kinds of help.} The model often took up what it was
given: computed notes changed its next step, and it mentioned cues nearly every time.
Its play improved at VC33, where the cue named the colour the recorded play had missed.
These experiments counted uptake and completion, not whether the model tested a rule
before relying on it, and the whole-game package, which included action probes, brought
no detectable gain. Section~\ref{sec:why} examines that step, from a noticed fact to a
revised rule and a test of it, in the plays that stopped.

%% file: sections/05-why-plays-stop.tex
\section{Hypothesis testing: diagnosis and training}
\label{sec:why}

We now ask whether the model compares its candidate rules with what they predict,
using one LS20 example and 109 selected plays that stopped on a level, and then whether
training it to check helps.

\subsection{Executable predictions and measurements}
\label{sec:critique}

On LS20 level~3, the fourth published play prints seventeen measurements of the move
bar: eight suggest that a move costs four cells, while three contradict this. It
nevertheless builds ``the bar = (value set at the last ring) - 4*(steps since the last
ring)'' into a route search, which predicts that a 45-action plan ends with eight cells left, but the game ends at action~22. The next reply first says the bar was not empty, then
reads it as empty in the final frame. The error could lie in the rule, in where it
applies, or in the measurement (rows and code in Appendix~\ref{app:trace-details}).

Checks of such rules in code are rare. In the 100 published plays, 145 of 9,170 programs, in 44
plays, define a function predicting the next state, and six of them compare its predictions with
recorded transitions. In 300 plays of our entry, 164 of 2,051 programs that call a
model-written step function also compare it with the record in the same program. These counts miss checks elsewhere and say nothing about their validity or effect. Unlike simulators validated by replay \citep{zhao2026memento3,zeng2026schema},
many model functions update self-parsed positions, so a check must also cover the
parser.

\subsection{The stopped plays, read against the games' source}
\label{sec:census}

A \emph{candidate causal rule} is a claim, right or wrong, about what an action changes
\citep{woodward2023sep}. A \emph{test} is an action or a look at the record undertaken
because of a prediction and followed by a comparison with it. A test is \emph{adequate}
if its outcome could refute the rule. A check is \emph{partial or nondiscriminating} if
it covers only part of the rule or would come out the same under the alternatives the model entertained.

A play \emph{stops} on a level when it ends there without completing it. In five
full-length runs of our entry, plays stopped 274 times on 69 levels that another play
completed. Two readings selected 109 stopped plays on 58 levels (56 of these and two
that no play completed): 59 at twelve high-loss levels, then 50 at the other 46. The
readers, frontier-model agents with a written brief, the game's source code and an
engine replay, assigned each play a primary failure category and any additional ones. A script confirmed every cited quotation, which establishes provenance only. Appendix~\ref{app:census} gives selection and examples. The findings describe
this selected sample.
\begin{figure}[ht]
\centering
\resizebox{\textwidth}{!}{\input{figures/stopped-plays-causes-bars}}
\caption{Where the 109 selected stopped plays went wrong, as judged afterwards: \textcolor[RGB]{145,145,145}{\rule{.9em}{.55em}}\,harness, \textcolor[RGB]{181,52,45}{\rule{.9em}{.55em}}\,model, \textcolor[RGB]{31,78,121}{\rule{.9em}{.55em}}\,scheduler or end of run. One primary category per play, plus any overlapping additional ones.}
\label{fig:causes}
\end{figure}
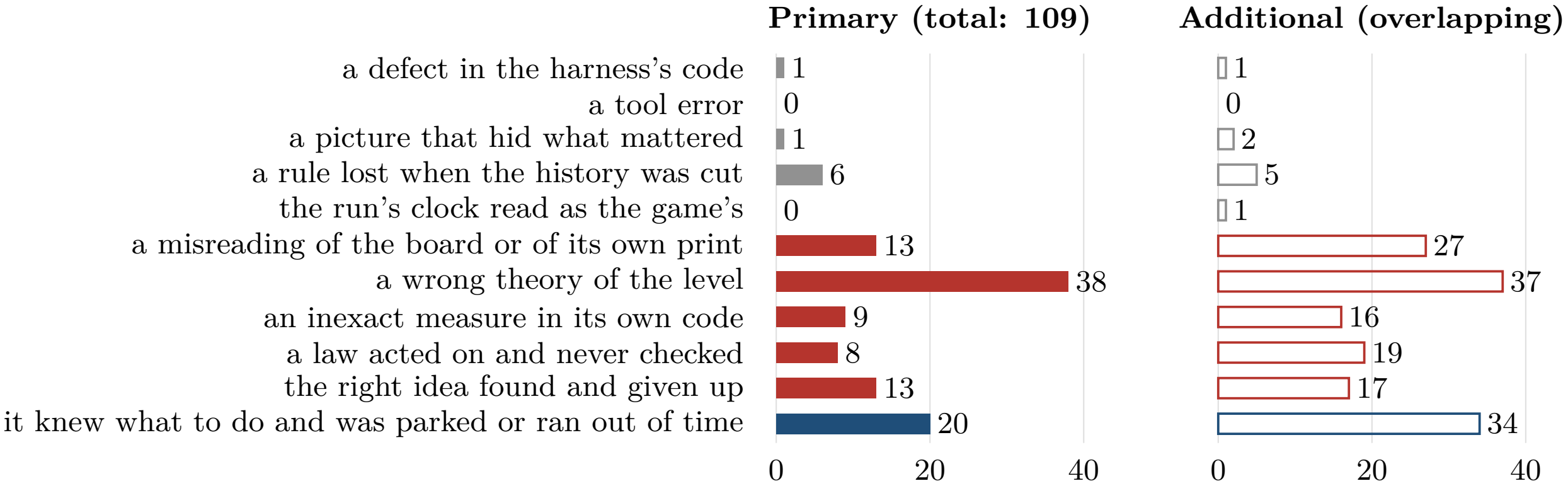

The primary failure lay in the model's behaviour in 81 plays, in the harness in eight,
and in scheduling or the time limit in 20. A wrong theory of the level was the largest category (38), and six harness cases were rules lost when the history was cut. Categories locate where a play went wrong without naming a sole cause, and a play
with a wrong theory may still have tested it.
\begin{figure}[ht]
\centering
\includegraphics[trim=0 36bp 0 0,clip,width=\textwidth]{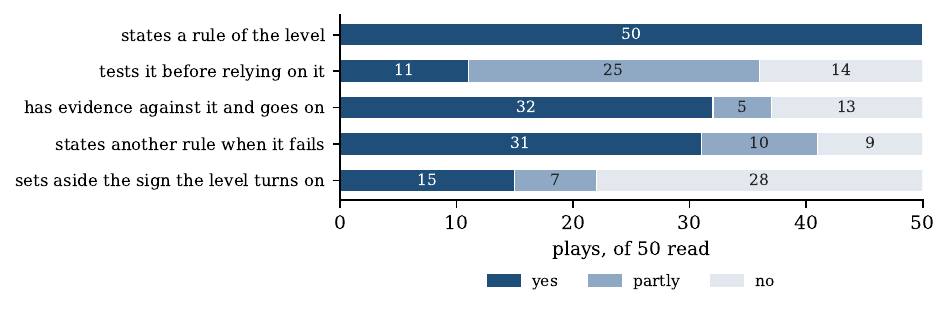}
\caption{Inquiry in the 50 stopped plays of the second reading: \textcolor[RGB]{31,78,121}{\rule{0.9em}{0.55em}}\,yes, \textcolor[RGB]{143,169,196}{\rule{0.9em}{0.55em}}\,partly, \textcolor[RGB]{227,232,238}{\rule{0.9em}{0.55em}}\,no. A partial check covers only part of a rule or cannot tell apart the alternatives the model was considering.}
\label{fig:inquiry}
\end{figure}

All 50 plays of the second reading stated a candidate rule. Eleven tested it before
relying on it, 25 in part, and 14 not at all. A CD82 play, for example, took completing
a level as validating its table of pours, though the completing plan used no left pour.
A blind rereading of 32 of these plays with the same brief found a test in 21 where this
reading found 8 (kappa 0.31), so how often plays test depends on the reader.
Thirty-two plays continued despite contrary evidence in their context, and 31 stated a new rule after a failure. Fifteen dismissed a relevant feature as decoration, animation or
display, and seven more did so in part. These labels can overlap, and a new rule need
not be correct or acted on.

For momentum, these counts suggest indicators of whether an unresolved relation stays central to later probes. They trace no causal chain from an event and measure no
drive. Successful plays show
what a discriminating action looks like: all six successful CN04 level~2 plays clicked
another piece within nine actions, whereas a stopped play concluded from one key press that only one piece moved, and
never clicked.

For harness design, the diagnoses suggest keeping learned rules across history cuts,
marking the harness's refusals as distinct from game responses, and validating
measurement and simulator code.

\input{sections/05-verification-training}

%% file: figures/stopped-plays-causes-bars.tex
\begin{tikzpicture}[x=1cm,y=1cm,font=\footnotesize]
\definecolor{causeharness}{RGB}{145,145,145}
\definecolor{causemodel}{RGB}{181,52,45}
\definecolor{causescheduler}{RGB}{31,78,121}

\node[font=\footnotesize\bfseries] at (9.6,.5) {Primary (total: 109)};
\node[font=\footnotesize\bfseries] at (14.2,.5) {Additional (overlapping)};
\foreach \base in {8,12.6} {
 \foreach \v in {0,20,40} {
  \draw[black!12] ({\base+.08*\v},.15) -- ({\base+.08*\v},-3.9);
  \node[anchor=north] at ({\base+.08*\v},-3.95) {\v};
 }
}
\foreach \label/\primary/\additional/\tone [count=\row from 0] in {
 {a defect in the harness's code}/1/1/causeharness,
 {a tool error}/0/0/causeharness,
 {a picture that hid what mattered}/1/2/causeharness,
 {a rule lost when the history was cut}/6/5/causeharness,
 {the run's clock read as the game's}/0/1/causeharness,
 {a misreading of the board or of its own print}/13/27/causemodel,
 {a wrong theory of the level}/38/37/causemodel,
 {an inexact measure in its own code}/9/16/causemodel,
 {a law acted on and never checked}/8/19/causemodel,
 {the right idea found and given up}/13/17/causemodel,
 {it knew what to do and was parked or ran out of time}/20/34/causescheduler} {
 \pgfmathsetmacro{\yy}{-.37*\row}
 \node[anchor=east] at (7.8,\yy) {\label};
 \ifnum\primary>0
  \fill[\tone] (8,{\yy-.105}) rectangle ({8+.08*\primary},{\yy+.105});
 \fi
 \node[anchor=west,inner sep=2pt] at ({8+.08*\primary},\yy) {\primary};
 \ifnum\additional>0
  \draw[\tone,line width=.7pt] (12.6,{\yy-.105}) rectangle ({12.6+.08*\additional},{\yy+.105});
 \fi
 \node[anchor=west,inner sep=2pt] at ({12.6+.08*\additional},\yy) {\additional};
}
\end{tikzpicture}

%% file: sections/05-verification-training.tex
\subsection{A preliminary verification-training experiment}
\label{sec:training}

Automatic labels on 14,119 replies from 116 plays of
58 community games marked two target behaviours: comparing a step function's predictions
with recorded transitions in the same program, and answering a surprise by comparing
expectation with observation and distinguishing explanations. Following
\citet{yeEtAl2026harnessZero}, feedback asked the model to rewrite replies to include these checks. A filter kept 736 of 832 rewrites, paired with their original contexts
without the feedback. We trained a low-rank adapter on them for one epoch and merged it
into the quantised weights.

Both models played the same 30 held-out games once in the same notebook
(Table~\ref{tab:rewrites}).
\begin{table}[ht]
\centering
\compacttable
\caption{One pass per model on 30 held-out games, at 48 minutes per game. In the first
two rows, counts give eligible replies followed by those meeting the checking criterion.}
\label{tab:rewrites}
\small
\begin{tabular}{lrr}
\toprule
 & unchanged & trained \\
\midrule
replies that define a law, of which checked against the record & 68, 12 & 678, 559 \\
surprise replies, of which compare and test expectations & 75, 4 & 597, 42 \\
surprises reported & 75 & 655 \\
generated tokens a reply, median & 378 & 1,370 \\
replies a play, median & 39 & 94 \\
levels, games won & 141, 22 & 55, 7 \\
\bottomrule
\end{tabular}
\end{table}

Among replies that define a rule, the share checking it against the record rose from
18\% to 82\%, while completed levels fell from 141 to 55 (Table~\ref{tab:rewrites}).
Reply length, the number of rule-defining replies and other behaviours changed too, so the result shows only that the model checked more often. It shows neither better
verification on the same opportunities nor that checking caused the decline.

Every training example contained a check, so the data could not teach when one is worth its cost. The open question is whether a check separates rival accounts and changes
later action (Appendix~\ref{app:training-details}).

%% file: sections/08-related-work.tex
\section{Related work}
\label{sec:related}

\textbf{Harnesses and world models.} Our cue extends Duck's published harness
\citep{bessisEtAl2026duck,franzen2026milestone2} (Section~\ref{sec:hidden}). Other
ARC-AGI-3 systems validate
executable world models \citep{zeng2026schema,lehmann2026tycho,rodionov2026ablation,
courtis2026opine,zhao2026memento3,wu2026kepler}, and earlier systems generate transition
programs under different task assumptions \citep{tangKeyEllis2024worldcoder,
daineseEtAl2024codeWorldModels,ahmedEtAl2025theorycoder,piriyakulkijEtAl2025poeworld}.

\textbf{Verification and evaluation.} Self-correction studies distinguish revision by
the model alone from correction with external support \citep{huang2024selfcorrect,
tyen2024mistakes,stechly2024selfverification,kambhampati2024llmmodulo}. ARC studies
compare verification variants and evaluate harness revisions on held-out games
\citep{rodionov2026ablation,lehmannVahdati2026harnessImprovement}. Our continuations
follow a proposed fix past the immediate response to completion within budget, compared with unaided recovery. Work on bounded reasoning likewise measures computation alongside
outcomes \citep{huang2026pace,lee2026flyby}.

\textbf{Inquiry and learning.} Peirce's account of inquiry
\citep{Pei31,Pei98,Sho07,bellucciPietarinen2023peircesAbduction}, studies of
confirmation bias, and theory-based game learning motivate discriminating actions
\citep{jhaveriEtAl2026failingToFalsify,tsividisEtAl2026theoryBasedLearning,
yiuEtAl2026empowerment}. Epistemic actions are taken to aid cognition \citep{kirsh1994epistemic}. Momentum concerns whether an unresolved relation keeps
organizing such actions in later play, even when the goal can be reached. Harness-Zero
distils scaffolded behaviour into a model \citep{yeEtAl2026harnessZero}, and process
rewards target abductive reasoning \citep{salimiEtAl2026cedarGRPO}. Our training comparison shows that more checking and more completion can come apart
(Appendix~\ref{app:related}).

%% file: sections/11-conclusion.tex
\section{Conclusion}
\label{sec:conclusion}

Reaching a goal does not show that the rule that led there is right. Our findings
support three practices for evaluating agents: (i) compare a proposed fix with unaided recovery from the same point,
(ii) ask whether a check can tell competing accounts apart and whether it informs later
action, and (iii) measure completion within budget, separately from uptake of
evidence or a corrected next step.

The LS20 comparison suggests \emph{momentum in inquiry} as an engineering target: keep
relevant unresolved relations as priorities for discriminating actions, even after local success. A drive to understand is our proposed, unestablished explanation of the
human--model difference. The intervention and
training results qualify it: extra facts or checks must prove their worth in later
play.

A future, open-ended evaluation could let learners pose their own questions, return to
unresolved relations after local success, and face situations in which rival accounts
call for different actions. It would ask whether self-directed probes yield corrections
that carry over as goals or mechanics change. Our evidence is limited to one
model and harness, selected decision points, interpretive annotations and one training comparison, and it shows neither a general
lack of drive nor a working policy for when to verify.

%% file: sections/A-what-the-model-is-sent.tex
\section{Harness and prompt specification}
\label{app:sent}

\input{sections/A-how-a-turn-runs}

\subsection{Prompt guidance and evidence cues}

The baseline system prompt is the published harness prompt
\citep{franzen2026milestone2}. The r28 differences are specified above. The exact
prompt and opener transcripts used here were extracted by script from the record of
r28's LS20 validation play on Kaggle, 12 October 2026. The full system prompt and two
opener transcripts are given in the file \texttt{prompt-and-openers.txt} that
accompanies this paper. The following
verbatim excerpts specify guidance relevant to the main behavioral analysis. They are excerpts and do not replace the full
prompt:

\begin{Verbatim}[fontsize=\footnotesize]
Carry forward supported knowledge as a starting hypothesis, while re-checking
anything contradicted by new evidence.

Use `current_frame.segmentation` as your primary view of the board -- objects, colors,
containment, adjacency, and cross-frame object hashes.

Optimize for the shortest reliable sequence that advances the current goal. If
confidence is low, program a discriminating probe and revise your approach from the
result.

After every action, verify whether gameplay objects changed or whether only a timer,
progress bar, or remaining-step bar moved. Do not treat HUD-only changes as evidence that
the move worked.
\end{Verbatim}

The prompt also advises treating edge bars as timers unless evidence connects them
to mechanics, checking \texttt{SPACE}/\texttt{UNDO} effects, inspecting local crops
rather than printing whole boards, and using search once the mechanics are understood.
Thus the model lacks game-specific rules, but receives explicit general testing guidance.

Openers repeat the available interfaces and retained function signatures without
certifying those functions as correct. Section~\ref{sec:hidden} supplies the exact
VC33 cue used in the continuation example, which includes both a factual description
and a pointer into the retained record.

\subsection{Submitted notebook and study configurations}
\label{app:configurations}

The submitted notebook is
\href{https://www.kaggle.com/code/andreicaioanei/semiotic-scaffold-r41}{semiotic-scaffold-r41},
Kaggle submission \texttt{57727775}. Its reported changes are verbatim reasoning
retention, the between-picture colour cue, a retrained speculative draft head,
and the startup keep-alive. The leaderboard result in the introduction describes this final entry, while the experiments concern the configurations and source material below.

\begin{table}[ht]
\centering
\compacttable
\caption{Source material and configurations supporting the reported studies.
Earlier releases and experimental variants remain distinct from the final r41 entry.}
\label{tab:study-configurations}
\small
\begin{tabularx}{\textwidth}{lL}
\toprule
Study & Source/configuration \\
\midrule
Baseline analysis (Section~\ref{sec:alone}) &
100 published plays from Franzen's harness, four on each of 25 games \\
LS20 model panels (Section~\ref{sec:momentum}) &
r28, calls 12 and 13, with a later correction at call 21 \\
Harness comparisons (Section~\ref{sec:help}) &
Experiment-specific intervention arms, with serving/configuration differences in
Appendix~\ref{app:harness-comparisons} \\
Stopped-play census (Section~\ref{sec:census}) &
Full-length entry runs. Retained examples identify r11/r28 runs and transcript
locations in Table~\ref{tab:censuswhere} \\
Training comparison (Section~\ref{sec:training}) &
Unchanged and full-scale adapter-merged models, each on the same 30 held-out games \\
\bottomrule
\end{tabularx}
\end{table}

%% file: sections/A-how-a-turn-runs.tex
\subsection{Runtime, observations, and entry configuration}
\label{app:loop}

\paragraph{Requests and execution.} A request contains the system prompt, retained
conversation, and opener. An ordinary opener reports the last program's actions and
shows the current board and changed cells at ten pixels per cell. First-level and
post-yield openers show the board alone, and post-death openers include the reset board,
death frame, and fatal change. Sampling uses temperature 0.7, reasoning enabled, and
at most 12,288 generated tokens per reply. A reply without a tool call receives a
nudge. Calls run in order until one acts. A turn ends after actions or when its
generated tokens reach 2,048, checked between requests. An inspection-only turn opens
again on the same board.

Each Python call runs in a fresh process for at most 30 seconds, with permitted
imports and output capped at 12,288 characters. Only eligible function definitions
persist throughout the game. The sole tool accepts Python code, in which \texttt{action(actions)} executes an ordered list drawn from \texttt{valid\_actions}
(directional controls, \texttt{SPACE}, \texttt{MOUSE}, or \texttt{UNDO} when offered).
\texttt{current\_frame} and \texttt{previous\_frame}
expose boards as letters, one per colour, together with metadata.
\texttt{history[i].frame} is the board after that entry's action, so \texttt{history[-1].frame} is the current
board itself.
\texttt{transitions} retains explicit before/after pairs. Intermediate boards are
retained even when absent from the opener images. The percept parses four-connected,
same-colour patches, with rotation-invariant shape/colour signatures, orientation,
size, boundary, enclosure, and contact. \texttt{frame\_diff} reports inferred movement,
rotation, appearance, disappearance, colour, and size changes. Component matches need not establish persistent object identity. The last action's animation frames and
diff timeline are separately inspectable.

\paragraph{Action guards and memory.} A batch stops after an action with no change
outside the four-cell board border. From level 2, another action call in the same
program is then refused, as are further calls after death, completion, or a win. Known-death and repeated-state guards are off. Automatic reset is a separate
transition and does not overwrite the fatal call's result. The conversation budget
is 118,272 estimated tokens including the prompt. When it is exceeded, the oldest exchanges are dropped until half
remains, and beyond 150 replies, 120 are retained. The game record is separate.

\paragraph{Serving and scheduling.} A fork of SGLang serves Qwen3.8-Flash-Next with
4-bit weights, an 8-bit cache, 139,264-token context, ten concurrent requests, and a
three-token speculative draft. From r28, a retrained draft head uses rejection
sampling. All games start together with 532-minute clocks and share ten service
slots, averaging about 48 minutes per game. A history cut releases a slot, and the parked play's priority depends on level weight, remaining levels, and actions/tokens spent.
To avoid the gateway's 15-minute idle closure during startup, the first game is reset
every five minutes until serving begins.

\paragraph{Entry differences.} Relative to the published harness, r28 has four
switchable changes: \path{ARC3_REASONING_VERBATIM} retains reasoning exactly,
\path{ARC3_TIMELINE_KEYS} preserves timeline fields without animation,
\path{ARC3_IMPORT_TIME} and \path{ARC3_ENTRY_AS_FRAME} permit clock imports and entry-as-frame access, and \path{ARC3_BETWEEN_PICTURES} adds the cue of
Section~\ref{sec:hidden}. Thus serving changes and observation interventions are
separate configuration choices.

\paragraph{Competition score.} Only completed levels score: the squared human/model
action-count ratio is capped at 1.15 and level $k$ has weight $k$
\citep{arcprize2026methodology}. Competition mode scores every game, permits one
scorecard/start per game, and makes reset restart the current level
\citep{arcprize2026competitionMode}.

%% file: sections/B-stopped-plays.tex
\section{Stopped-play selection and trace provenance}
\label{app:census}

\paragraph{Selection and annotation.}
The first reading (17 October 2026) selected 59 stopped plays at twelve levels:
highest-loss levels, three early levels usually completed by the entry, and the
two largest never-completed levels. The second (20 October) selected one stopped
play at each of the other 46 levels, plus a second at the four with most stops,
giving 50 plays. Together these cover 109 stops at 58 levels, 56 completed by
another play and two never completed. These cases were selected on purpose and do not form a random sample of stopped plays.

Frontier-model readers used a written brief and each game's source, and recorded actions were replayed on the game's engine. Primary and additional categories were assigned
against that record, and a script checked quotation provenance. This is not independent
validation of the annotations. The testing criteria are defined in
Section~\ref{sec:census}. The first reading and its counting script, the brief, the
selection, and the combined census are kept in the authors' repository and are
available on request.

\paragraph{A blind rereading.} The testing question is whether, before
the play relied on its rule for a sequence of actions, it ran a check whose outcome
could have refuted the rule. A deliberate single probe whose result it compared with
what the rule predicted counts, and so does code that compares the rule's predictions
with recorded transitions. Acting on the rule and reading the result afterwards does
not, unless the action was chosen as a test.
Readings found a rule stated in all 32 plays, and contrary evidence in context in 21
and 19 of them. They agreed on contrary evidence in 24 of
32 and on a feature set aside in 24. They agreed on
the primary category in 16. On testing before reliance they agreed in
17 and all 15 disagreements went upward. Seven partial checks and six
plays without a check became tests, and two plays without a check became partial
checks, so the reading found 21 tests, 10 partial checks and one play without a check
where the second reading found 8, 15 and 9. It credited announced probes, such as
``Let me test: press LEFT and see what happens to green'', simulations that reproduced
the record, and in one play a move whose success would have won the level. In three of
the six plays that rose from no check, the second reading had quoted nothing. The
readers applied one criterion with different thresholds, so the share of plays that
test before reliance is a reading, about one in four by the second reading and two in
three by the rereading.

\begin{table}[ht]
\centering
\compacttable
\caption{Counts underlying Figure~\ref{fig:causes}, separated by reading. Each play has one primary category, and additional categories may overlap.}
\label{tab:census}
\small
\begin{tabularx}{\textwidth}{Lrrrr}
\toprule
Category & 17 Oct & 20 Oct & Primary & Additional \\
\midrule
Harness: code defect & 1 & 0 & 1 & 1 \\
Harness: tool error & 0 & 0 & 0 & 0 \\
Harness: picture hid relevant evidence & 0 & 1 & 1 & 2 \\
Harness: rule lost after truncation & 4 & 2 & 6 & 5 \\
Harness: run clock read as game clock & 0 & 0 & 0 & 1 \\
Model: misread board or printed output & 6 & 7 & 13 & 27 \\
Model: wrong theory of the level & 21 & 17 & 38 & 37 \\
Model: inexact measurement in code & 4 & 5 & 9 & 16 \\
Model: rule acted on without a check & 4 & 4 & 8 & 19 \\
Model: right idea found and abandoned & 6 & 7 & 13 & 17 \\
Scheduler/run end: plan available, play stopped & 13 & 7 & 20 & 34 \\
\midrule
Plays & 59 & 50 & 109 & \\
\bottomrule
\end{tabularx}
\end{table}

\paragraph{Representative distinctions.}
The following cases support the diagnoses in Section~\ref{sec:census}. They illustrate the labels and do not establish isolated
causes. Table~\ref{tab:censuswhere} locates the
quoted material in the transcripts.

\emph{Candidate rule versus discriminating action.} On CN04 level~2, a stopped play
inferred ``Only the white plug moves. So the white plug is the player-controlled
object'' from one key press and never clicked. All six successful plays clicked
another piece within nine actions. A different play proposed a click test, but clicked
background and concluded ``MOUSE on a solid yellow cell = no-op. So MOUSE doesn't
select plugs.'' Proposing a test and executing a valid comparison are distinct.

\emph{Stated test versus executed action.} On M0R0 level~1, a simulator prohibited
squares from overlapping although overlap is required. The play wrote ``In my sim
I disallowed that. What actually happens?'' and proposed pressing RIGHT, but its next actions began with DOWN. The stated probe was not the executed one.

\emph{Testing versus correct measurement.} Another CN04 search omitted a piece's
offset, found ``total all-mated configs: 0'', and rejected the correct goal. Restoring
the offset yields four solutions. Executing a search does not validate its measurement.

\emph{Context loss and harness responses.} On RE86 level~4, a play stated ``So the
mechanic: passing over a colored square recolors the shape.'' After history cuts,
it called the level~7 goal ``color-agnostic''. On SU15 level~4, the harness refused a
second UNDO after an unchanged board, and the play read this as ``UNDO is a no-op''.
These cases distinguish retained game knowledge and interface behavior from game rules.

\emph{Nondiscriminating validation.} On CD82 level~4, a play justified its pour table
with ``These were validated by level 3's completion'', although its successful level~3
plan used no left pour. Completion therefore supplied no test of the left-pour entry.

\begin{table}[ht]
\centering
\compacttable
\caption{Transcript locations for the retained examples. Run names identify full-length entry runs, and play is the pass of the game. Line numbers locate the quoted statement or
its immediate context.}
\label{tab:censuswhere}
\small
\begin{tabularx}{\textwidth}{lclrlL}
\toprule
Game & Level & Run & Play & Lines & Statement/context \\
\midrule
CN04 & 2 & fullR28b & 2 & 2663 & Only the white plug moves \\
CN04 & 2 & fullR11 & 3 & 3394, 3609 & Proposed click and no-op conclusion \\
M0R0 & 1 & fullR28 & 1 & 3186 & Simulator forbids overlap, proposed probe \\
CN04 & 2 & fullR28 & 0 & 8417, 8600 & Zero configurations, rejected goal \\
RE86 & 4, 7 & fullR11 & 2 & 6968, 17376 & Earlier dye rule, later colour-agnostic claim \\
SU15 & 4 & fullR28b & 3 & 9638 & UNDO interpreted as no-op \\
CD82 & 4 & fullR28 & 0 & 13515 & Completion used as validation \\
\bottomrule
\end{tabularx}
\end{table}

%% file: sections/C-related-work-in-full.tex
\section{Scope of the evaluation comparison}
\label{app:related}

The closest distinctions behind Section~\ref{sec:related} concern what an evaluation
holds fixed and which outcome it measures. Schema checks executable models against
recorded transitions and verifies execution, and Rodionov compares world-model,
simplification, and verification variants \citep{zeng2026schema,rodionov2026ablation}.
These assess system designs or components. Our continuations instead test a proposed
repair from one recorded state and preceding context, with unchanged recovery as
the comparison and generated tokens as the matched budget. They estimate a local completion effect and make no claim about a benchmark-wide
policy advantage.

Tyen and colleagues separate locating an error from correcting it
\citep{tyen2024mistakes}. Our next-response probes similarly assess a targeted
behavior, whereas the continuation endpoint asks whether a repair survives subsequent
play. A note, a multi-component whole-game package, and a sparse cue are therefore different treatments and cannot serve as interchangeable evidence for one
mechanism.

The Peircean vocabulary organizes observable proposals, predictions, and comparisons.
The human--model panels motivate momentum in inquiry. They establish neither the presence nor the absence of an
internal drive. Thus the paper combines a local intervention
evaluation with a proposed account to be tested, and the account does not serve to validate the
intervention results.

%% file: sections/D-further-measurements.tex
\section{Supporting experimental details}
\label{app:measures}

This appendix supplies the comparison, trace, and training details used in
Sections~\ref{sec:alone}--\ref{sec:why}.

\subsection{Harness comparisons and cue deployment}
\label{app:harness-comparisons}

\paragraph{Relational summaries.} Hosted-endpoint runs on six public games at
25 minutes each completed 25, 26, and 24 levels with summaries versus 24 and 23 without, and two simultaneously started pairs yielded 26 versus 27 and 23 versus 23.
The competition-GPU comparisons are in Section~\ref{sec:telling}.

\paragraph{Correction-package configurations.} Four hidden-game entries carrying
correction-related components scored 28.73, 23.01, 27.20 and 20.38, and four without scored 28.73, 31.53, 22.24, and 30.48. One further run per arm scored zero without
available Kaggle logs, so its failure cause is unconfirmed. Serving configurations
also differed: these scores do not isolate correction effects. The public-game
comparison used the configurations that scored 31.53 without and 20.38 with the
package. Summaries were retired on 4 October and correction packages on 9 October.

\paragraph{Cue trigger and census.} The deployed rule inserts at most one cue per
transient colour per level. Offline replay inserted 48 cues in 4,510 openers across
28 plays and 11 games. Manual inspection classified these as 28 selection/held-piece
marks, five click-switched parts, five VC33 sockets, five tiles revealed on stepping
off them, and five edge marks or overlays.

\paragraph{Competition score context.} The first cue-bearing entry scored 30.48 and 28.18, whereas the same package without the cue had scored 22.24. Entry r28, with a
faster server, scored 33.01 and 24.66. These runs document the submission and do not
identify a hidden-game cue effect.

\subsection{Human action intervals and model reply lengths}
\label{sec:people}

The seven recordings replayed in Section~\ref{sec:alone} are drawn from eighteen
completed LS20 trajectories collected in workshops at Leibniz University Hannover,
with an experimenter present, no game-specific instructions, and timestamped inputs
\citep{ABA+26}. These are eighteen trajectory files, and the number of distinct participants is not
verified. The trajectories took 405--1,111 actions (mean 655). Re-run on the game's
engine, the seven recordings in the current game version reproduce the recorded board
after every action in five, after all but ten actions in a sixth, and after the first
404 actions in the seventh. The four model plays are reproduced in full.

The human measure is the interval after an action with more than three changed cells
in the key region, compared with other actions of the same level. Its medians are 0.96 and
0.46 seconds, and each trajectory has a longer median after key changes, by a factor
of 1.4--3.2. For the model, the same board criterion selects the first reply after a turn:
four full-length baseline
plays and nine 25-minute entry plays. Median reply lengths are 1,168 versus 1,027
tokens (ratio 1.14). Eight of thirteen plays have longer replies after changes. Ratios range from 0.49 to 2.48, with 0.80, 1.91, 1.13, and 0.94 in the four
full-length plays.

Human elapsed time and model reply length are different quantities. Model turns can contain several actions, and the samples differ in completion and duration.
The comparison describes responses to key changes. It does not show equivalent
deliberation or concept formation, and it does not directly measure the proposed
internal drive.

\paragraph{The first turn of the key.} The counts in Section~\ref{sec:momentum} follow
rules written before they were read. A turn on level~1 is an action other than a reset
after which the key's orientation differs and the lives are unchanged. A repeat is a
later turn in the same life, before the level is completed. An interval is the time to
the next action in the same session, kept when positive and at most 30 seconds. Eleven
of the eighteen files log an older build of the game, whose first level starts the
player five cells to the right and which the engine cannot replay. Turns were therefore
read from the recorded boards, as more than three changed cells in the key region with
the lives unchanged. This reading found the engine's turns action for action in all
eleven plays both could read, seven human and four model. Every trajectory turned the
key on level~1, and the game's white frames around the key and the goal showed at that
first turn in all eighteen. The interval after it exceeded the trajectory's median after
the level's other actions in 17 of 18, by a median ratio of 3.20, and its median was
1.77 seconds. Nine trajectories turned the key again in the same life, eight of them two
actions later, off the rotator and back on, the soonest possible, and one three actions
later. They were two of the seven trajectories in the current build and seven of the
eleven in the older one. The model plays are the four full-length LS20 plays of our earlier entry r28, which differ from the published plays of Table~\ref{tab:ls20people}. All four turned the
key and drew the white frames, and one turned it again ten actions later, with nine
turns in that life. Four plays support this count and no comparison of rates.

\subsection{Code carried between levels}
\label{app:carry}

In 92 competition-model plays, most at half the competition time, later levels
frequently called functions written earlier. In the first 20,000 generated tokens
of a level, 71 levels that subsequently remained uncompleted had a median 27\% of
programs calling earlier functions, versus 19\% in 60 completed levels. New functions
per 10,000 tokens were 0.9 versus 1.5, and programs calling earlier computing functions
(search or rule implementations) were 13\% versus 5\%. These are associations between code reuse and later completion. They neither measure
whether every reused rule was correct nor show that reuse caused failure.

\subsection{LS20 trace measurements and route search}
\label{app:trace-details}

The rows below are printed estimates from the LS20 example in Section~\ref{sec:critique}. They were not checked against the engine's state.
Interpreting the discrepancy requires action/frame alignment and checking the
measurement itself.
\begin{Verbatim}[fontsize=\small]
(149, 'DOWN', 16, 8, (45, 54))
(150, 'DOWN', 8, 8, (45, 54))
(151, 'UP', 8, 8, (40, 54))
\end{Verbatim}
The next reply states ``the bar = (value set at the last ring) - 4*(steps since
the last ring)'' and implements it in the route search:
\begin{Verbatim}[fontsize=\small]
RINGVAL={1:92,2:84}
            nt=t-4
            if np_ in (YA,YB) and nr<2: n_nr=nr+1; nt=RINGVAL[n_nr]
            if nt<MINBAR: continue
\end{Verbatim}

\subsection{Verification-training checks}
\label{app:training-details}

Section~\ref{sec:training} gives the source corpus, one-epoch adapter procedure,
and held-out comparison. The target marks were written before training. In the
rewrite source corpus, the replay pattern occurred in 20/504 replies calling a step function, and the surprise-response pattern occurred in a few dozen replies.
Of 832 requested rewrites, 736 passed the pattern filter. Manual readings of
40 rewrites per kind agreed with it on 40 and 36, respectively. These checks confirm that the target patterns are present. They do not establish that
individual checks were correct or helped play.

The main evaluation uses the full-scale merged adapter and one pass on the same
30 held-out games as the unchanged model, at 48 minutes per game.

\input{sections/D-momentum-panels}
\input{sections/D-intervention-details}

%% file: sections/D-momentum-panels.tex
\subsection{LS20 level progression}
\label{app:momentum-panels}

Table~\ref{tab:ls20levels} records the game-source progression used in
Section~\ref{sec:alone}. The human--model panels and their interpretive relation
maps remain in the main paper (Section~\ref{sec:momentum}).

\begin{table}[ht]
    \centering
\compacttable
\caption{ What each level of LS20 adds, and the number of actions the competition scores
the level against. }
    \label{tab:ls20levels}
    \begin{tabularx}{\textwidth}{clL}
    \toprule
    Level & Actions & New on the level \\
    \midrule
    1 & 22 & the key, a goal with its figure, a tile that turns the key, a bar of moves
      that costs a life when empty, a sign that lights when the key matches \\
    2 & 123 & two tiles that refill the bar, twice the cost of a move, three turns needed,
      no sign \\
    3 & 73 & a tile that changes the key's colour, two strips that throw the player to the
      next wall \\
    4 & 84 & a tile that changes the key's shape \\
    5 & 96 & a changing tile that moves along a track with every action \\
    6 & 192 & two goals, and all three changing tiles move along tracks \\
    7 & 186 & fog that hides the board beyond twenty cells from the player \\
    \bottomrule
    \end{tabularx}
\end{table}

%% file: sections/D-intervention-details.tex
\subsection{Correction probes and continuation details}
\label{app:intervention-details}

\paragraph{Wording-sensitive probes.} A computed six-row-shift description was
taken up in 15/16 samples versus 1/16 without it. Two computed descriptions of a
level-ending action were taken up in 0/32 samples, whereas a human-written correction
addressing the model's claim was taken up in 10/16. These are local behavioral
responses, distinct from reference-step agreement and task completion.

\paragraph{Continuations and cue variants.} Each arm samples ten continuations
from the same recorded board and preceding context. Twenty continuations generated
approximately 0.6--0.9 million tokens and took 20--35 minutes on the competition GPU.
Round~1 enabled both colour and earlier-animation cues and round~2 the colour cue alone, so Table~\ref{tab:rounds} separates their results.
For the LS20 earlier-animation cue, the model encountered the colour tile before
trying the goal with a wrong key in 8/20 continuations in each arm. With the cue,
all 20 first replies mentioned teleporting, wrapping, or sliding, and without it 19/20 did so by the third reply. The cue was therefore noticed without improving
this prespecified behavioral target.

\paragraph{Response scoring.} Colour-cue mention, board retrieval before the first
action, and completion are separate endpoints. The retrieval mark required the
model's program to hold the referenced history entry before acting: round~1 met it
in 19/50 treated versus 2/50 unchanged continuations, below the prespecified target
of 25/50. The cue itself supplies factual content, which can be used without
retrieving that board. Successful-run token/action medians in Table~\ref{tab:rounds}
exclude unsuccessful continuations and do not measure overall efficiency.